\documentclass[conference]{IEEEtran}
\IEEEoverridecommandlockouts
\usepackage{cite}
\usepackage{amsmath,amssymb,amsfonts}
\usepackage{algorithmic}
\usepackage{graphicx}
\usepackage{textcomp}
\usepackage{xcolor}
\usepackage{hyperref}
\def\BibTeX{{\rm B\kern-.05em{\sc i\kern-.025em b}\kern-.08em
    T\kern-.1667em\lower.7ex\hbox{E}\kern-.125emX}}
\begin{document}

\title{Reinforcement Learning for Inventory Control under Regime-Switching Demand: A Controlled Comparison of LSTM, MLP, and (s,S) Policies} 

\author{\IEEEauthorblockN{Alex Haines, Nathanael L. Baisa}

\thanks{Alex Haines and Nathanael L. Baisa are with the School of Computer Science and Informatics, De Montfort University, Leicester LE1 9BH, UK. Email: nathanael.baisa@dmu.ac.uk.}

}


\maketitle

\begin{abstract}
Classical inventory policies such as (s,S) are provably optimal under stationary demand but cannot adapt when demand switches between hidden regimes, and memoryless deep reinforcement learning policies, confined to a fixed observation window, cannot maintain a belief over latent demand structure. We ask whether a recurrent policy offers a measurable advantage over both a memoryless policy and a classical baseline under regime-switching demand, and how that advantage scales with supply-chain lead time. We build a \textit{custom} Gymnasium environment in which daily demand is generated by a two-state hidden Markov process with Poisson emissions, so that the regime is known at evaluation time but hidden from the agent. An optimised (s,S) heuristic, an MLP PPO agent, and an LSTM Recurrent PPO agent are trained under matched compute and evaluated at lead times of 0, 3, 6, and 10 days over 30 deterministic episodes each. The recurrent policy outperforms both alternatives at every lead time, and its margin over the MLP widens monotonically with lead time, from parity at lead time 0 to 2.17$\times$ at lead time 10. Principal component analysis of the LSTM hidden state explains the widening gap: the representation shifts from encoding a binary regime indicator at short lead times to encoding multi-step pipeline state at long lead times. Recurrent memory becomes the decisive architectural factor as the credit-assignment horizon lengthens. The source code is available at \textcolor{red}{\url{https://github.com/nathanlem1/RL-InventoryControl}}.
\end{abstract}


\section{Introduction}
Real-world demand rarely follows the stationary distributions assumed by classical inventory theory~\cite{RobJorWil22}. A product can sit in a quiet baseline for months before a promotion, a competitor exit, or a macroeconomic shift moves it into a different regime, and the drivers of these shifts are usually hidden and cannot be predicted in advance. A policy tuned to a single stationary process is no longer optimal once the underlying regime changes, and the cost of that miscalibration grows with the length of the replenishment pipeline: decisions taken under the wrong regime continue to land as stock arrivals long after the regime itself has moved on.

Deep reinforcement learning (RL)~\cite{Murphy25} is a natural candidate for adaptive inventory control because a learned policy can condition on richer state than a reorder point allows. In practice, however, much of the literature either trains and evaluates under stationary demand, so the question of regime adaptation never arises, or trains on real demand traces, in which the regime structure is unobserved and the contribution of any architectural choice to regime inference cannot be isolated. Recurrent architectures such as the Long Short-Term Memory (LSTM)~\cite{HocSch97} are appealing here because their hidden state can accumulate evidence across many timesteps, but whether that capacity is actually used to track latent demand structure, and whether it translates into control performance, has not been cleanly tested. A controlled environment in which the regime sequence is known at evaluation time but hidden from the agent is needed to answer the question on its own terms.

We construct such an environment and study three policies across four lead times. We make three contributions. First, a like-for-like quantitative comparison showing that a recurrent policy outperforms a memoryless one under regime-switching demand and that the gap widens with lead time. Second, a mechanistic account of why, by projecting the LSTM hidden state into two dimensions and showing that its representation shifts from a regime indicator at short lead times to a pipeline-tracking state at long lead times. Third, a decomposition against an optimised classical baseline in which the gap between (s,S) and the Multilayer Perceptron (MLP) measures the value of state-dependent control and the gap between the MLP and the LSTM measures the additional value of recurrent memory.
Four research questions (RQs) guide the study. RQ1: does an LSTM policy outperform a memoryless MLP policy under regime-switching demand? RQ2: does any advantage grow as lead time increases (lt = 0, 3, 6, 10)? RQ3: can either RL approach beat an optimally tuned (s,S) baseline under non-stationary demand? RQ4: to what extent can the LSTM infer and track the latent regime, relative to the limited demand history available to the MLP?

\section{Related Work}

Inventory control balances stock availability against cost. Scarf~\cite{Herbert60} proved that an (s,S) reorder rule, which replenishes to level S whenever the inventory position falls to or below s, is optimal under stationary Markovian demand with fixed ordering costs, and it remains the standard benchmark. Demand is frequently non-stationary; however, seasonality, macroeconomic change, and sudden virality produce structural shifts in which the optimal policy itself changes, and a fixed (s,S) rule cannot adapt, over-ordering in low-demand periods and stocking out in high-demand ones.
Hamilton~\cite{James89} formalised Markov regime-switching for economic time series, capturing structural breaks that fixed-process AutoRegressive Integrated Moving Average (ARIMA) models cannot. Hidden Markov models (HMMs) generalise the idea: a latent Markov chain drives transitions between states, each with its own emission distribution, with Rabiner~\cite{Rabiner89} the standard reference. A two-state HMM with Poisson emissions is a natural model of alternating high- and low-demand regimes, with a transition matrix governing persistence.
Under a hidden regime, inventory control is a partially observable Markov decision process (POMDP), in which optimal decisions depend on a belief state built from observation history rather than the current observation alone~\cite{LesMicAnt98}. An (s,S) rule cannot represent this latent structure, and a memoryless MLP cannot carry information beyond a short observation window. An LSTM~\cite{HocSch97} can accumulate evidence in its hidden state and so approximate belief-state updates; its gated constant-error carousel mitigates vanishing gradients and supports long-range dependencies.
RL frames inventory control as a sequential decision problem. Tabular Q-learning~\cite{WatDay92} is infeasible in large state spaces; deep Q-network
(DQN)~\cite{VolKorDav15} scales via function approximation but is less stable than the on-policy Proximal Policy Optimization (PPO)~\cite{JohFilPra17}, whose clipped surrogate objective constrains updates and is valuable in noisy-reward settings. Boute et al.~\cite{RobJorWil22} survey deep RL for inventory, and Gijsbrechts et al.~\cite{GijBouVan22} show RL can beat classical heuristics in some stationary-demand settings, but most work assumes fixed demand or smooth seasonality, leaving hidden-regime inference underexplored. Transformer policies such as the Decision Transformer~\cite{CheLuRaj21} have fixed context windows ill-suited to arbitrarily long horizons, making the LSTM the practical choice here. The closest empirical study, Hachaichi et al.~\cite{HacCheAff20}, compares PPO against deep deterministic policy gradient (DDPG) under stationary demand with a fixed observation window; it leaves open how learned policies fare against a competitively tuned heuristic, whether recurrence extracts value a fixed window cannot, and how either performs under regime-switching demand. We address these three questions directly.

\section{Method}

\subsection{Environment and Demand Process}
We study a single-product, lost-sales inventory problem. Demand is generated by a two-state HMM with Poisson emissions: a low-demand regime ($\lambda = 5$) and a high-demand regime ($\lambda = 25$) alternate under a symmetric transition matrix with P(stay) = 0.97, giving a mean persistence of about 33 steps and a long-run mean of $E[\lambda] = 15$ units per step. This persistence is long enough that a five-step demand window cannot reliably identify the current regime, yet short enough that each 5,000-step episode contains many transitions. At each step the environment samples the next latent state and draws Poisson demand from it; the ground-truth regime is logged for evaluation but never exposed to the agent.
The environment subclasses \textit{gymnasium.Env}~\cite{TowKwiBal25} for use with Stable-Baselines3~\cite{AntAshAda21}. Each step represents a day and each episode runs 5,000 steps. Within a step the chosen action is placed into the pipeline, any orders whose lead time has elapsed arrive into on-hand inventory, demand is realised, sales are settled with unmet demand recorded as lost sales, and the reward is computed. A hard cap of 500 units on the total inventory position bounds the state space, and reset() randomises initial on-hand inventory (10–80 units) and each pipeline slot (0–20 units) so that learned policies must generalise across starting states.

\subsection{State, Action, and Reward}
At the longest lead time the observation is a 19-dimensional vector: on-hand inventory, the most recent demand, the total inventory position, the pipeline utilisation ratio, the full pipeline vector of length equal to the lead time, and the last five observed demands. The five-step window gives the MLP partial access to recent trends without making it memory-capable, and because it also strengthens the MLP it ensures the comparison reflects recurrence rather than differential information access. All features are normalised and clipped to [0, 2] in a Box observation space.
The action is a discrete order quantity from the bins {0, 5, 10, 15, 20, 25, 30, 40}, coarse enough to keep the output layer small but fine enough not to bottleneck the policy. The per-step reward is $tanh((revenue - total\_cost)/50)$, which squashes reward into (-1, 1) and prevents large magnitudes from destabilising PPO advantage estimates while preserving gradient. The cost structure reflects a representative retail setting: a selling price of £2.00 per unit, an order cost of £0.40 per unit, a holding cost of £0.30 per unit per day, a stockout penalty of £1.00 per unit, and a fixed order cost of £0.20 per non-zero order.

\subsection{Policies}
The classical baseline orders up to S when the inventory position falls to or below s and orders nothing otherwise, with the integer quantity rounded to the nearest action bin. The baseline is held to the same maximum order and discrete bins as the learned policies, and the optimal (s, S) 
is found per lead time using a grid search, with the search window centred on a \textit{base number}, which for each lead time is \textit{lead time multiplied by $E[\lambda]$}.
The memoryless agent is \textit{stable\_baselines3.PPO} with an MlpPolicy of two 128-unit hidden layers in actor and critic; it sees only the current observation and must infer the regime from the embedded five-step window. The recurrent agent is \textit{sb3\_contrib.RecurrentPPO} with an MlpLstmPolicy: a single 128-unit LSTM feeds a shared 128-unit head, and the hidden and cell states are carried across timesteps within an episode and reset only at episode boundaries. The MLP is given two hidden layers so its parameter count approximately matches the LSTM, ensuring any gap reflects recurrent memory rather than capacity.

\subsection{Training and Evaluation}
Both agents are trained with PPO configured identically wherever possible (Table~\ref{tab:config}). Each rollout collects 10,000 transitions, more than one full episode, giving clean Monte-Carlo return targets; updates run for six epochs over 2,000-sample minibatches. Each model is trained for 5,000,000 timesteps under matched compute. Before RL, both policy and value heads are warm-started by behavioural cloning on 50,000 transitions from an (s,S) teacher (s = 81, S = 91) for five epochs, minimising cross-entropy on the teacher's actions plus mean-squared error on Monte-Carlo returns. For the LSTM, cloning uses zero-initialised hidden states, so only the feedforward heads are warm-started and the recurrent dynamics are learned entirely by PPO.
Each policy is evaluated over 30 deterministic episodes per lead time on a fixed seeded set (seeds 42–71), so every policy faces an identical demand and regime sequence. We report mean episodic reward together with service level (units sold over units demanded), stockout fraction, mean on-hand inventory, and total orders. A deterministic evaluation runs every 200,000 steps and the best-deterministic-reward snapshot is retained for the final evaluation, guarding against late-stage collapse. Pairwise two-sided Welch's t-tests compare each policy pair at every lead time.

\begin{table}[h]
  \begin{center}
    \caption{Key environment and training configuration (shared across both agents except where noted).}
    \label{tab:config}
    \begin{tabular}{|l|c|r|}
     \hline
     Parameter & Value \\
     \hline
     Demand regimes  & $\lambda~in~\{5, 25\}$ (Poisson) \\
     Transition matrix & P(stay) = 0.97, P(switch) = 0.03 \\
     Episode length	& 5,000 steps \\
     Inventory-position cap	& 500 units  \\
     Action set	& \{0, 5, 10, 15, 20, 25, 30, 40\} \\
     Costs (price/order/hold/stockout/fixed) &	2.00 / 0.40 / 0.30 / 1.00 / 0.20 \\
     Reward	& tanh((revenue - cost)/50) \\
     Total timesteps / rollout	& 5,000,000 / 10,000 \\
     Minibatch / epochs	& 2,000 / 6 \\
     Learning rate, $\gamma$, GAE $\lambda$ & 1e-4, 0.99, 0.90 \\
     Clip range, entropy coef. & 0.1, 0.05 \\
     LSTM / MLP architecture &	1×128 LSTM / 2×128 MLP \\
     Evaluation	& 30 deterministic episodes, \\
                & \ seeds 42–71  \\
     \hline
    \end{tabular}
  \end{center}
\end{table}
\noindent

\section{Experimental Results}

\subsection{Reward Across Lead Times}
\begin{figure}[h]
\begin{center}
  \includegraphics[width=1.0\linewidth]{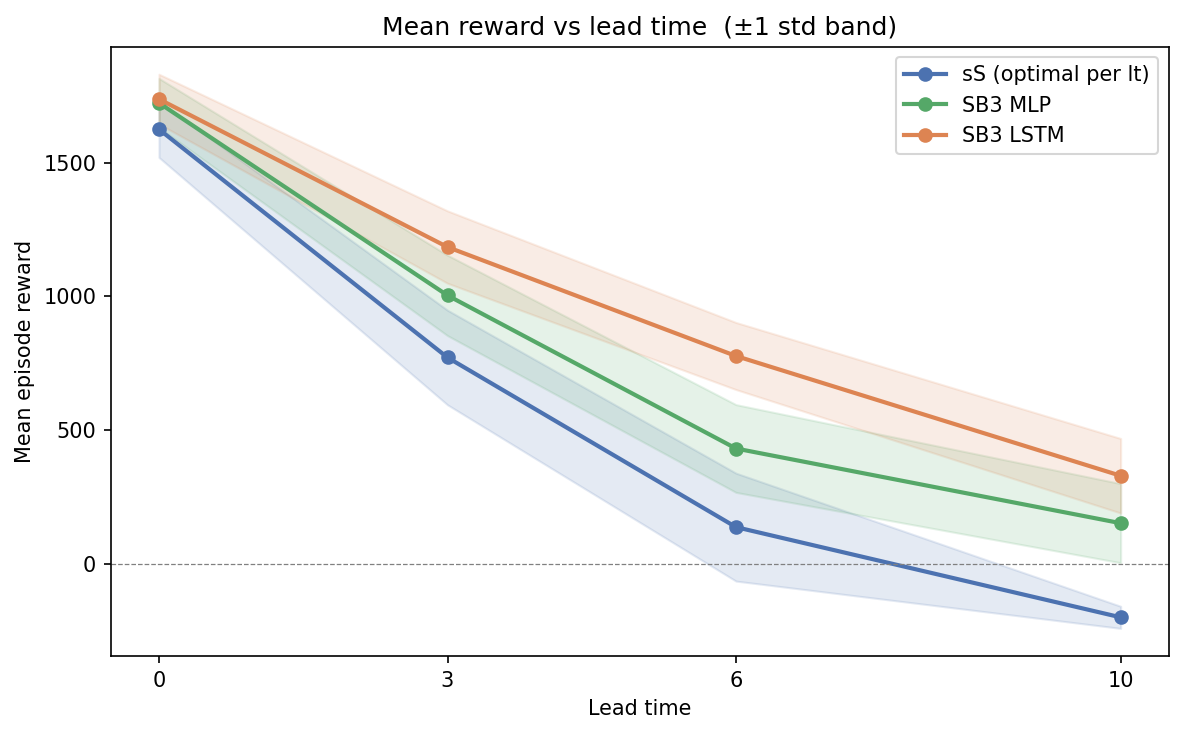} \\  
\end{center}
   \caption{Mean evaluation reward versus lead time for the LSTM, MLP, and (s,S) policies, with one-standard-deviation (std) bands over 30 episodes.}
\label{fig:comparison}
\end{figure}

Table~\ref{tab:resultstab1} and Figure~\ref{fig:comparison} report the headline results. Three findings stand out. First, the LSTM is the strongest policy at every lead time, and its margin over the MLP grows with lead time: the two tie at load time (lt) = 0 (a 1.01$\times$ ratio) before the LSTM pulls ahead to 1.18$\times$, 1.80$\times$, and 2.17$\times$ at lt = 3, 6, and 10. Second, both learned policies dominate the (s,S) optimum at every setting, and the gap widens sharply with lead time; at lt = 10 the (s,S) optimum is negative (about -201) while the LSTM still earns +329, so the heuristic cannot operate profitably in a regime where the LSTM can. Third, although standard deviations (std) rise with lead time they remain proportionally similar across architectures, indicating stable ordering across episodes.

\begin{table*}[h]
  \begin{center}
    \caption{Mean episodic reward (+/- std) per policy and lead time over 30 deterministic episodes (seeds 42–71), with optimal (s,S) parameters.}
    \label{tab:resultstab1}
    \begin{tabular}{|l|c|c|c|c|r|}
     \hline
     Lead time (lt) & LSTM (RecPPO)	& MLP (PPO)	& (s,S) optimum	& (s,S) params\\
     \hline
     0	& 1737.9 +/- 93.2 &	1723.5 +/- 92.0	& 1624.9 +/- 105.3	& s=15, S=30 \\
     3	& 1184.5 +/- 135.4	& 1003.6 +/- 150.5	& 771.2 +/- 176.9	& s=85, S=95  \\
     6	& 776.8 +/- 125.5	& 430.7 +/- 164.2	& 136.6 +/- 201.9	& s=140, S=150  \\
     10	& 328.9 +/- 139.5	& 151.4 +/- 148.0	& -200.7 +/- 41.3	& s=65, S=75  \\
     \hline
    \end{tabular}
  \end{center}
\end{table*}
\noindent

\noindent

A clear inventory-service trade-off emerges as lead time grows. At lt = 6 the LSTM earns the highest reward (776.8) despite the lowest service level (0.84), carrying mean inventory of 27.4 units against the MLP's 42.9 and the (s,S) rule's 55.9: it deliberately accepts some stockout exposure for materially lower holding cost. The same pattern holds at lt = 10 (LSTM 328.9 at service 0.74; MLP 151.4 at 0.76). Lacking persistent memory, the MLP cannot distinguish a transient demand spike from a regime shift and defaults to a conservative buffer, whereas the LSTM achieves higher reward while ordering less, indicating that it times replenishment rather than stockpiling defensively.

\subsection{Statistical Significance}
To establish whether the reward differences observed across the 30-episode evaluation are statistically significant, pairwise two-sided Welch's t-tests were performed between each policy pair at every lead time (Table~\ref{tab:resultstab2}). At lt = 0, both learned policies significantly outperform the (s,S) baseline (LSTM: t = 4.76, $p < 0.001$; MLP: t = 4.21, $p < 0.001$), but the LSTM-MLP gap is not significant ($t = 0.59$, $p = 0.555$), consistent with the finding that recurrent memory provides no real advantage when replenishment is immediate. From lt = 3 onwards every pairwise comparison is significant at $p < 0.001$. The LSTM-(s,S) statistic rises with lead time (t = 12.77, 16.91, 19.47 at lt = 3, 6, 10), as does the MLP-(s,S) statistic (t = 7.17, 7.76, 12.09), confirming that the gap between learned policies and the heuristic widens as the replenishment pipeline grows. The LSTM-MLP comparison is significant at every non-zero lead time, with the largest separation at lt = 6 (t = 9.02) where the recurrent advantage in regime inference is most pronounced; the smaller statistics at lt = 3 (t = 4.81) and lt = 10 (t = 4.70) reflect the higher episode-to-episode variance at those settings, not a smaller mean difference. 

\begin{table}[h]
  \begin{center}
    \caption{Welch t-statistics for pairwise reward differences at each lead time (30 episodes).}
    \label{tab:resultstab2}
    \begin{tabular}{|l|c|c|r|}
     \hline
     Lead time (lt) & LSTM vs (s,S)	& MLP vs (s,S) \\
     \hline
     0	& 4.76 	& 4.21 \\
     3	& 12.77	& 7.17 \\
     6	& 16.91	& 7.76 \\
     10	& 19.47	& 12.09 \\
     \hline
    \end{tabular}
  \end{center}
\end{table}
\noindent

\subsection{Training Behaviour}
Both architectures trained stably: approximate Kullback-Leibler (KL) divergence stayed in 0.002–0.004, policy-gradient loss was near zero, and there was no gradient explosion. The LSTM's clip fraction (about 0.05–0.15) exceeded the MLP's (about 0.01–0.05), consistent with a more actively adapting policy, and critic explained variance stabilised near 0.6 for both, a useful but loose baseline given the regime-switching stochasticity. Several runs had not converged at 5M steps: the LSTM at lt = 3 was still improving at the final checkpoint, the MLP at lt = 10 was the least converged of all eight runs, and the LSTM at lt = 10 peaked near 4.4M before overshooting toward a near-zero-stockout policy that over-accumulated pipeline inventory. In every case the best-deterministic snapshot was retained, so the absolute gaps should be read as values at matched but possibly limited compute.

\subsection{Hidden-state Analysis}
\begin{figure*}[h]
\begin{center}
  \includegraphics[width=1.0\linewidth]{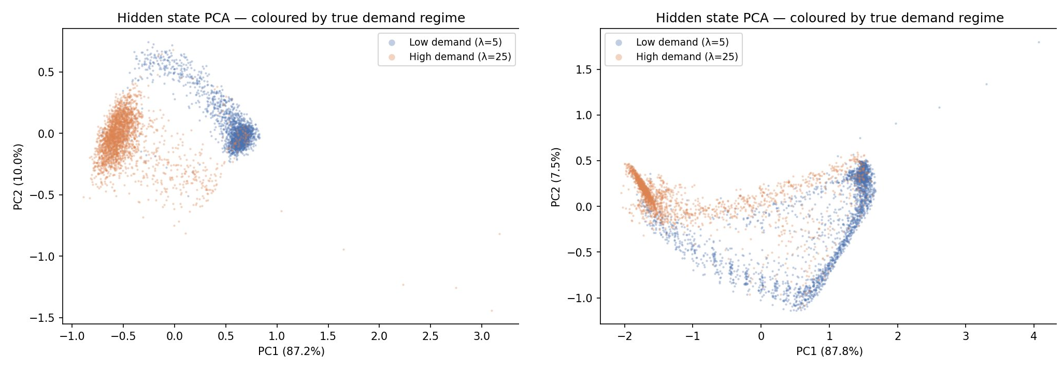} \\  
\end{center}
   \caption{PCA of the LSTM hidden state coloured by true demand regime. (Left) At lt = 0 the regimes form two separable clusters along PC1. (Right) At lt = 10 they interleave in a V-shape, indicating a shift from regime-indicator to pipeline-tracking encoding.}
\label{fig:hiddenstatePCA}
\end{figure*}

To address RQ4 we project the LSTM actor's 128-dimensional hidden state onto its two leading principal components over a 5,000-step deterministic episode. The geometry changes systematically with lead time (Figure~\ref{fig:hiddenstatePCA}). At lt = 0, the two regime clusters overlap but separate along PC1 (which captures 87.2\% of variance); PC1 switches sharply at each regime change and on-hand inventory never exceeds about 18 units, so the hidden state encodes a compressed reactive control variable. As lead time grows the clusters separate more cleanly (PC1 93.6\% at lt = 3), the low-demand cluster fans into an arc at lt = 6 with the trajectory tracing a curved loop, and at lt = 10 the projection forms a V-shape whose two arms correspond to inventory building as deliveries arrive and inventory being drawn down by demand.

\noindent

Regime separation is weakest at lt = 10, with high- and low-demand points interleaved throughout the V, while inventory shows its sharpest gradient across the two arms (one arm roughly 80–140 units, the other near zero). The progression reads as a shift in what the hidden state is for: a regime indicator at short lead times and a pipeline-tracking state at long lead times. This is precisely the representational change one would predict if the LSTM were solving the credit-assignment problem by inferring latent demand structure from observation history, and it is the direct, interpretable evidence RQ4 calls for. The MLP, confined to a fixed window, cannot reproduce this capacity.

\section{Discussion}
The central result, that recurrence ties the MLP at lt = 0 and more than doubles its reward by lt = 10, matches the theoretical motivation: when the optimal action at time t depends on observations from t - k, a memoryless policy is structurally unable to recover the lost information while a recurrent policy can encode it in its hidden state. The decomposition is clean. The (s,S)-to-MLP gap is the value of state-dependent control, and the MLP-to-LSTM gap is the additional value of recurrent memory, which grows as the credit-assignment horizon lengthens. The trade-off the LSTM learns, higher reward at lower service level, is itself evidence of regime inference: it neither over- nor under-orders for long after a transition.
Several limitations qualify the findings. First, compute: several runs were unconverged at 5M timesteps, so the LSTM-MLP gap is best read as a value at matched but limited compute, and the relative ordering could shift with more training. Second, the demand model is a single two-state Poisson HMM with a fixed transition matrix, one slice of the regime-switching family; closer emission means, rarer or more frequent switching, more than two regimes, non-stationary transitions, or non-Poisson emissions would each be expected to move the gap. Third, the single-product, single-echelon, lost-sales structure with a fixed cost schedule is stylised. Finally, the capped action size is a structural ceiling that constrains the (s,S) rule more than the learned policies, and a single training seed per cell leaves run-to-run variance unreported.

\section{Conclusion}
We compared a recurrent PPO policy, a memoryless PPO policy, and an optimised (s,S) heuristic for inventory control under hidden-regime demand across four lead times. The LSTM outperforms the MLP at every lead time (lt), with parity at lt = 0 and a margin growing to 2.17$\times$ at lt = 10 (RQ1, RQ2); both learned policies beat the optimally tuned (s,S) at every lead time (RQ3); and PCA of the LSTM hidden state reveals a representational shift from regime indicator to pipeline-tracking state, the plausible mechanism for the widening gap and direct evidence of latent-regime inference (RQ4).
The most valuable extension is a higher-compute re-run (10–20M timesteps with a decaying entropy coefficient) to settle the unconverged runs and place the LSTM-MLP gap on firmer ground. Beyond that, systematic sweeps over emission separation, switching rarity, number of regimes, and emission distribution would map where recurrence stops adding value; multi-product and multi-echelon networks with stochastic lead times would test coordination across echelons; and a regime-conditional (s,S) baseline that switches between two (s,S) pairs given an offline HMM regime estimate would isolate the value of latent-state inference from that of end-to-end control.

\bibliographystyle{IEEEtran}
\bibliography{refs}

\end{document}